\documentclass[runningheads]{llncs}

\usepackage{amsmath}
\usepackage{amssymb}
\usepackage{amsfonts}
\usepackage{bm}
\usepackage{graphicx}
\usepackage{subfigure}

\begin{document}

\title{A Hybrid Federated Learning Based Ensemble Approach for Lung Disease Diagnosis Leveraging Fusion of SWIN Transformer and CNN}

\titlerunning{FL-Based Lung Disease with Fusion of SWIN Transformer and CNN}

\author{
Asif Hasan Chowdhury\inst{1}\orcidID{0000-0002-1066-0542} \and
Md. Fahim Islam\inst{2}\orcidID{0000-0002-7732-8237} \and
M Ragib Anjum Riad\inst{3}\orcidID{0000-0002-6500-8009} \and
Faiyaz Bin Hashem\inst{4}\orcidID{0000-0001-8819-8100} \and
Md Tanzim Reza\inst{5}\orcidID{0000-0001-8964-1565} \and
Md. Golam Rabiul Alam\inst{6}\orcidID{0000-0002-9054-7557}
}

\authorrunning{Chowdhury et al.}

\institute{
BRAC University, Dhaka-1212, Bangladesh\\
\email{\{asif.hasan.chowdhury, md.fahim.islam, m.ragib.anjum.riad, faiyaz.bin.hashem, tanzim.reza, rabiul.alam\}@bracu.ac.bd}
}

\maketitle

\begin{abstract}

The significant advancements in computational power create a vast opportunity
for using Artificial Intelligence in different applications of healthcare and medical
science. \textbf{A Hybrid FL-Enabled Ensemble Approach For Lung Disease Diagnosis Leveraging a Combination of SWIN Transformer and CNN} is
the combination of cutting-edge technology of AI and Federated Learning. Since,
medical specialists and hospitals will have shared data space, based on that data,
with the help of Artificial Intelligence and integration of federated learning, we can
introduce a secure and distributed system for medical data processing and create
an efficient and reliable system. The proposed hybrid model enables the detection
of COVID-19 and Pneumonia based on x-ray reports. We will use advanced and
the latest available technology offered by Tensorflow and Keras along with Microsoft-developed Vision Transformer, that can help to fight against the pandemic that the
world has to fight together as a united. We focused on using the latest available
CNN models (DenseNet201, Inception V3, VGG 19) and the Transformer model
SWIN Transformer in order to prepare our hybrid model that can provide a reliable
solution as a helping hand for the physician in the medical field. In this research, we
will discuss how the Federated learning-based Hybrid AI model can improve the
accuracy of disease diagnosis and severity prediction of a patient using the real-time
continual learning approach and how the integration of federated learning can ensure
hybrid model security and keep the authenticity of the information.

\end{abstract}
\section{Introduction}
Integrating artificial intelligence with medical science has created a new dimension
to the treatment world. Computer-assisted diagnosis can help doctors to sense any
forthcoming lethal diseases beforehand. Nowadays, doctors across the world tend to
rely more on AI as it is improving swiftly. We are looking to develop a system that can identify lung diseases that can help medical people during the treatment procedure. We are aware that we need to be very cautious to develop a system that will analyze patients' medical reports and identify the disease of patients. We are focusing to use the AI-driven approach to address the lung disease of patients. In this research, we proposed an ensemble method to detect lung diseases. We focus to achieve preferable accuracy with better performance therefore we build an ensemble
method for our research. We have ensembled several latest AI-based Algorithms like VGG-19, Inception V3, DenseNet201, and Vision Transformer developed by Microsoft. We have combined the outcome from this algorithm to develop a model that will be unique and reliable for lung disease detection. Furthermore, we took the help of Federated learning to ensure the data privacy of sensitive patient medical images. We want to build a network through federated learning where different hospitals will stay
connected together and share their effective treatment models. These effective models will be used to improve the performance of the central model which will be considered the core of the entire system. This global model will be updated based on the outcome from the local model through the federated learning-based network to ensure high security during the weight transfer process between the models that will be in different parts of the world.

\subsection{Research Objective}
The main objective of this research is to build a fusion model using transfer learning and a transformer model to save the patient from ARDS, a severe state of the lung. We aim to improve the outcomes of studies on existing transfer learning models by adding SWIN transformers to make a fusion model and also using federated learning we aim to ensure healthcare data security, low latency, and less power consumption.
\begin{itemize}
    \item Lung disease detection using the Deep CNN model to analyze the severe conditions of patients.
    \item Improve existing Transfer Learning models by building a new Fusion model that combines Transfer Learning and Transformer Learning.
    \item Integration of Shifted Window (SWIN) Transformer model for better accuracy and detection.
    \item Utilization of federated learning models for ensuring data security, low latency, less power consumption, and better accuracy.

\end{itemize}


\section{Literature Review}

Kassania et al. proposed a deep CNN approach to detect COVID-19 from X-ray and CT images [1]. The authors tried to get a better solution to the over-fitting issue in deep learning due to the small number of training images by using a transfer learning strategy. Firstly, Kassania et al. applied the image normalization technique to get better visual quality of input images. In the feature extraction step, the authors used a transfer learning strategy to lessen computational resources and accelerate the convergence of the network as their dataset is very limited. Finally, the authors developed a web-based application to assist doctors in detecting COVID-19 by uploading X-ray or CT images. This research contains some limitations such as few training data samples and security assurance. In our paper, we fed our model with a comparatively larger training set while ensuring the privacy of patients. We have developed a fusion model to get more accuracy in an efficient way.
\\

Hemdan et al. introduced a framework of deep learning named COVIDX-Net to diagnose COVID-19 from X-ray images [2]. The COVIDX-Net framework consists of seven DCNNs of different architectures and those are VGG19, DenseNet201, InceptionV3, ResNetV2, InceptionResNetV2, Xception, and MobileNetV2. The authors fed their models with a limited number of data. Hemdan et al. got better results using VGG19 and DenseNet201 whereas the result with InceptionV3 was not satisfactory. This paper shows a comparison among existing DCNN models with limited data.
\\

Minaee et al. [3] trained 4 state-of-the-art convolutional networks for COVID-19 detection. Jiang et al. proposed a model which is a combination of SWIN transformer and Transformer and Transformer to make a classification of COVID-19 from a dataset of X-ray images [4]. The existing conventional models have slow computational power and large sizes. Using the SWIN Transformer model can increase the computational speed with the size of the image. In this paper, Gu et al. proposed a fusion model that combines Swim transformer blocks and a lightweight U-Net type model that has an encoder-decoder structure [5].
\\

Li et al. have discussed the mechanism of federated learning for securing data and overcoming challenges [6]. First, the authors have mentioned the challenges one might face implementing federated learning. Such as expensive communication, system heterogeneity, statistical heterogeneity, and privacy concerns. Then, Li et al. come up with solutions to those challenges. For more communication efficiency, the authors have pointed to multiple methods like local updating, and decentralized training. As model training may differ for devices’ hardware specifications, the authors recommend using an asynchronous scheme that applies an optimization algorithm parallelly. Sometimes, data for federated learning are divided among devices non-identically, to resolve this problem meta-learning and multitask learning have been added to federated settings [7].
\\

A framework named MOCHA [8] is used for learning separate but related models for each device. Bonawitz et al. [9] introduced a protocol for individual model updates. In this protocol, the central model will not be able to see the local updates but will observe the aggregated result at the end of every round. This method is an inspiration for our proposed model where we will implement federated learning to overcome privacy leakage.

\section{Methodology}\label{SCM}

\begin{figure}[htbp]
\centering{\includegraphics[scale=0.088]{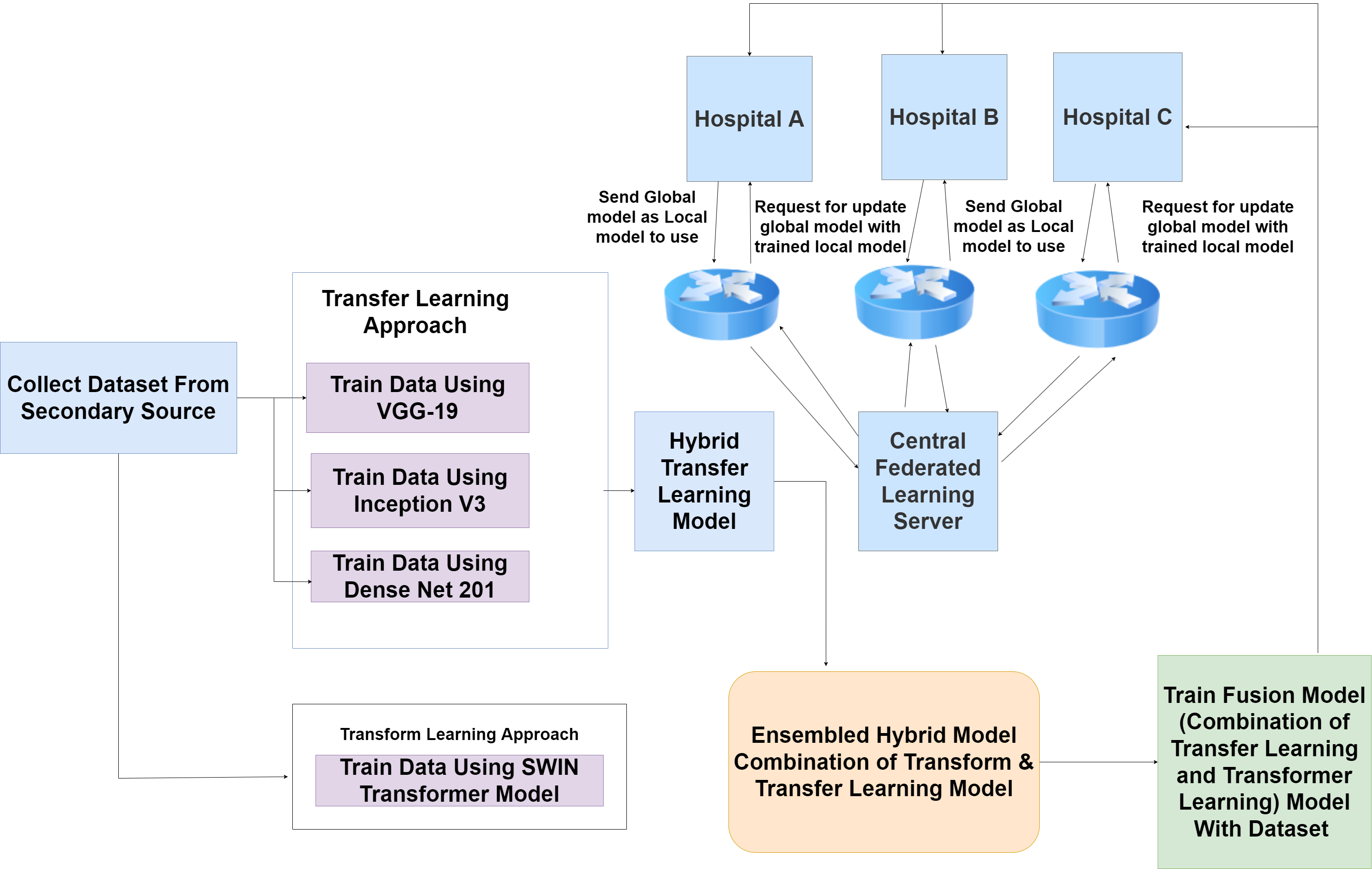}}
\caption{Top Level Over View of Proposed Disease Detection System}
\label{fig}
\end{figure}
In Figure 1, we have shown the top-level overview of the proposed lung disease detection system. First, we will collect
the data. Next, we will start to train the existing transfer-based learning models
using our dataset. We have planned to work with VGG-19, Inception V3, DenseNet201 Model. Then we will save the best-trained model individually and Ensemble
them together. Next, we will start to train the transformer-based model that we
decided to work with the SWIN transformer Model. We will also train this model
using our dataset.
Next, we will combine the trained SWIN transformer model with the Transfer
learning-based ensemble model to create our own hybrid model. Next, we will
train the hybrid model with our available dataset to complete our model training
and validation. Moreover, we are using a federated learning approach to secure
each individual model that will be held by the hospitals. Hospitals will share the
best finding outcome with the global server to ensure better accuracy and outcome.

\subsection{Dataset Collection Process}

Data collection is the most significant task to start building the CNN model. The
initial step of the work plan is to collect data from different primary sources.
We know medical data is sensitive and difficult to manage. We initially looked to
find medical data from different hospitals. In most cases, we were not able
to manage the same disease-related information. Then, we looked at different disease-related papers for the dataset. We get different datasets but still being open
source datasets some people alternated the large dataset with wrong files and corrupted files. Thus we have become careful enough during the selection process of
the dataset.

\begin{figure}[htbp]
\centering{\includegraphics[scale=0.15]{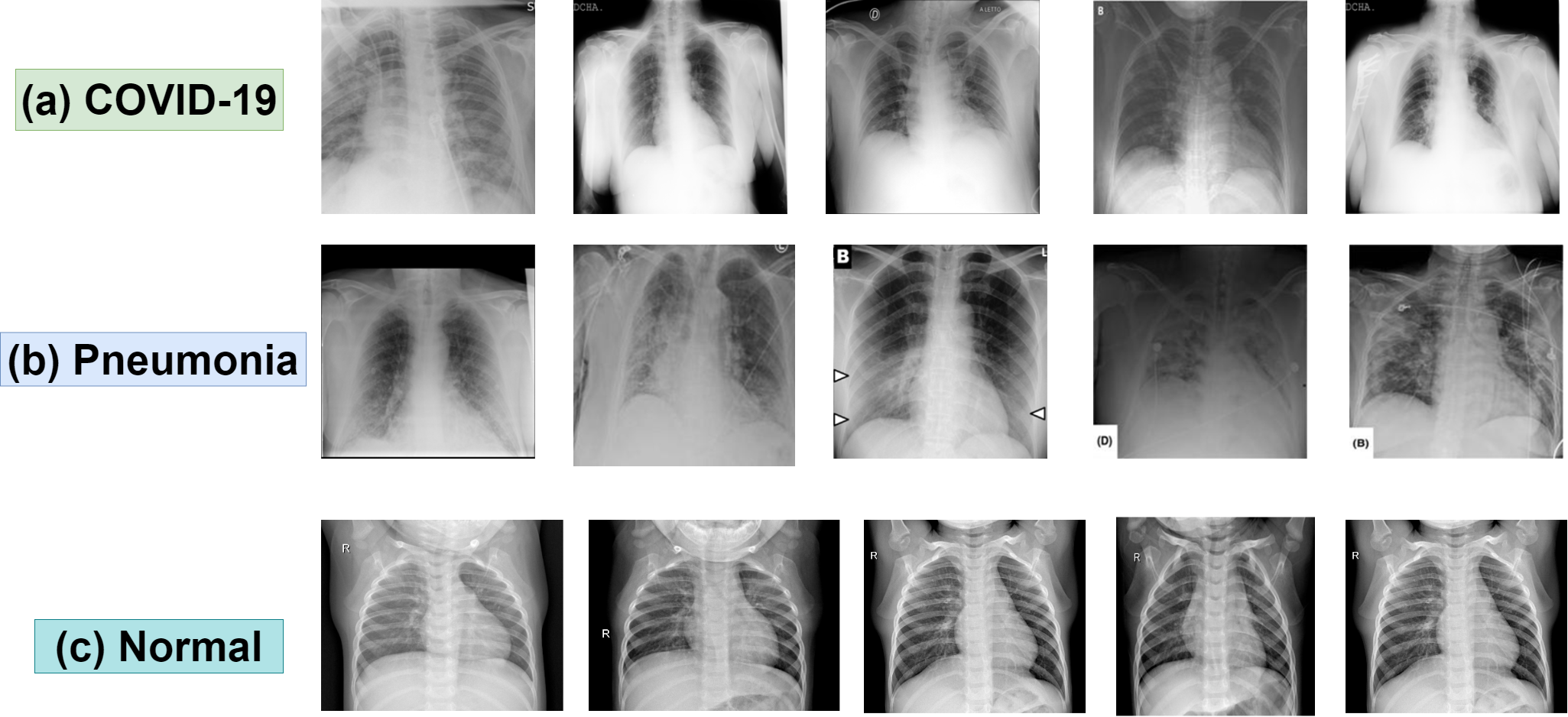}}

\caption{Sample Dataset of \break
  (a) COVID-19; (b) Pneumonia; (c) Normal}

\label{fig}
\end{figure}

\subsection{Dataset Analysis and Interpretation}
we processed the image data using the available pre-processing techniques. We have
identified corrupted images in the first place. Then, we removed the wrong image
data. For example, the X-Ray dataset contains CT-Image data. Then, we figure
out the number of images available for the training process and increase the data by
the data augmentation process. Next, we used re-scaling, mage-rotation, horizontal-rotation, and zoom-range for the image processing process.
We need to split the dataset into two categories known as training sets and testing
sets. The train set will do the task of training the dataset and preparing local models
for different hospitals. The test set will do the job of testing the predicted diseases.
We followed the convention of the training set(80\%) and testing set(20\%). We will
have more accurate results if we can increase the ratio of the training set.

\subsection{Classification and Decision Classifier}
As our paper requires multiple predictions, we implement the VGG-19, Inception-
v3, and DenseNet 201 for it. In addition, we have used a transformer-based learning
model that is the SWIN transformer Model.
These models will exhibit real-time predictions for each individual. The probabilistic results of the model will help patients and medical practitioners to detect
the disease.

\subsubsection{VGG-19}VGG 19 is the latest version of the Visual Geometry Group model series. This model series is the successor of the AlexNet This model consists of 19 layers. Out of 19 layers, 16 layers are Convolutional layers and 3 fully connected layers and 5 MaxPool layers, and 1 SoftMax layer.
\\
\\
In order to categorize the photos into 1000 object categories, Simonyan and Zisser-
man (2014) presented the VGG19. There are numerous 3 x 3 filters used by each
convolutional layer. Because each convolutional layer uses numerous 3 x 3 filters, it
is a highly well-liked technique for classifying images.
\subsubsection{Inception-v3}
The third generation of Inception convolutional neural network designs is known as Inception-v3. Among other improvements, the Inception-v3 convolutional neural network architecture makes use of Label Smoothing, Factorized 7 x 7 convolutions, and the addition of an auxiliary classifier to move label information lower down the network (along with the use of batch normalization for layers in the sidehead).\\ \\
The Inception architecture is built to function successfully even when memory and
compute resources are severely limited. Though the architectural simplicity of VG-
GNet is appealing, it comes with a considerable computational cost when assessing
the network. As Inception is lower while higher-performing successors, It is feasible
to utilize the e Inception networks in big-data scenarios.
\\ 

The architecture of an Inception-v3 is progressively built, step-by-step.
1. Factorized Convolutions: This decreases the number of parameters used in a
network, which lowers computational efficiency. It also monitors the effectiveness of
the system.
2. Smaller convolutions: This causes training to go more quickly by substituting
smaller convolutions for larger ones. Say a 5 X 5 filter has 25 parameters; replacing
it with two 3 X 3 filters results in only 18 (3*3 + 3*3) parameters [10].

\subsubsection{DenseNet201}A typical convolutional neural network is started with an input image and runs through the network to get a predicted label. The output of the previous convolutional layer is used by the subsequent convolutional layer, which receives the input image from the previous layer and constructs an output feature map.
\\ \\
But, In a DenseNet architecture, All layers are densely connected. That means
an inter-layer connection exists between each layer. Moreover, L connections exist
between L levels, one between each layer and the layer below it. So, There are
L(L+1) /2 direct links in the network. The feature maps of all layers before it are
utilized as inputs for each layer, and its own feature maps are used as inputs into all
levels after it [5]. The DenseNet architecture’s dense connectivity can be represented as:
\begin{align}
    x(l) &= H(l)([x(0),x(0),...,x(l-1)])
\end{align}
\subsubsection{SWIN Transformer}SWIN Transformer is stated as Shifted Windows Transformer. This is basically a hierarchical Transformer that is computed with shifted windows. To address the challenges of differences between two domains, such as large variations in the scale of visual entities and the high resolution of pixels in images compared to words in the text, this hierarchical Transformer or SWIN Transformer was proposed.
\\ \\
In SWIN Transformer architecture, first of all, it splits an RGB input image into non-
overlapping patches using modules like ViT(Vision Transformer), Each of the split
patches are called “token” which features are set as a concatenation of raw RGB
pixel values.

\subsection{Brief Work Steps}
This study seeks to aggregate locally trained models by retrieving them from local
servers. The centrally trained model will be sent to all nearby hospitals following
the implementation of Federated Learning.
Most hospitals don’t want to share their patient data for privacy issues. So, In our
research methodology, we do not take each hospital’s datasets. For those hospitals
that are willing to connect to this system, we give them a model that is already
trained with some test datasets, which is called a global model. This global
model will be sent to each connected hospital and give them access to fit and train their
dataset in that model to contribute to the improvement of accuracy. That model weight sent
to each hospital is called a local model. After successfully fitting and retraining local
model will be sent to a central server.
The server will take the top 80\% models based on their accuracy and test whether
the model is more accurate than the previous global model, and if the global model will
be overridden by the most accurate local model. The CNN and SWIN Transformation algorithms are the foundation of the model. After using the hybrid ensemble
CNN model with the VGG19, Inception-v3, and DenseNet algorithms, SWIN Transformation was included, and the primary model was developed.
Figure 4.1 is the whole proposed methodology of our research. First of all, X-Ray
Data is being collected to make a global model. The Dataset is being preprocessed
for training in our predictive model; we have to remove some corrupted image
data, also and an augmentation is being applied.

\subsection{Transfer and Transformer Fusion Model}
The main part of our model we used hybrid model V1 of VGG19, Inception-V3,
and DenseNet201. All the data is being trained separately in VGG19, InceptionV3, and DenseNet201. After the Ensemble process, the accuracy will increase and
make the model more reliable. Furthermore, adding SWIN-Transformer to the main
fusion model based on the transfer learning model and transformer model will have
the leverage to combine transfer learning and transformer-based learning, which will
ensure novel factors for the model.

\begin{figure}[htbp]
\centering{\includegraphics[scale=0.4]{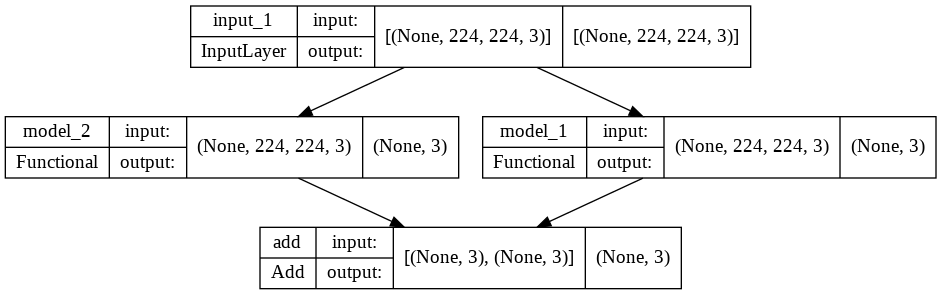}}
\caption{Transfer and Transformer Fusion Model}
\label{fig}
\end{figure}

\subsection{Federated Learning Centralized Server}
The main Hybrid trained model will be our initial global model of FL integrated
central server. Then the server will send the global to each hospital’s local devices.
After that, the hospitals will have their local model on their local devices and continue
to work with that. If any of the hospitals have enough datasets to train or fit again
in the local model they can fit into that and make a request to the central server
to update that model with the global model. The Central server will check and
take the top 80\% model with better accuracy and update the global model with
a local model whose model’s predictive performance is better than the previous global
and other local models. This work will be done in a loop whenever any hospital will
request to update the global model.

\section{Implementation and Result Analysis}

Disease Prediction is one of the most sophisticated examples of advanced computational ability. Now it is possible to analyze and detect diseases based on CT-Image
and X-Ray images. Thanks to the advancement of Artificial Intelligence. There are
several AI-based models that can do the job of prediction.
We have used some cutting-edge technology to predict the difference between COVID-19, Pneumonia, and normal X-Ray Images. We have used VGG-19, Inception V3,
DenseNet 201 and SWIN Transformer to create our model that can provide reliability to medical practitioners. We have tried to come up with the best approach
that can help the medical sector from our computer science field. Moreover, we compared different outcomes that helped us in the way of making the hybrid model more
advanced compared to the existing disease detection model with much reliability.
\subsection{Implementation}
\subsubsection{VGG19}
VGG-19 is the latest pre-trained model from VGG net architecture. It is the updated version of VGG-16. The size of each layer is now 47, which was 41 before in
VGG-16. Also, it has variants of filter sizes 64, 128, 256, and 512. In our VGG-19
model, we have added 3 batch normalization layers along with two dense layer sizes
of 128 and 64.
Moreover, we also made the middle layers trainable false so that we can avoid overfitting issues. Also, we added a dropout layer with a value of 0.5 to make sure that
our model is safe from the overfitting problem. In addition, we have used image
sizes (224, 224,3) for our overall processing steps. We have been careful during
the choice of size, considering our processing unit capability and required time to
complete the training without any issues.
In addition, during training time, we have calculated steps per epoch and the number
of epochs based on the available number of images that also keeps our model safe
from overtraining. Moreover, With all these careful steps, we have found 94.4
validation accuracy during our training time.

\begin{figure}[htbp]
\centering{\includegraphics[scale=0.5]{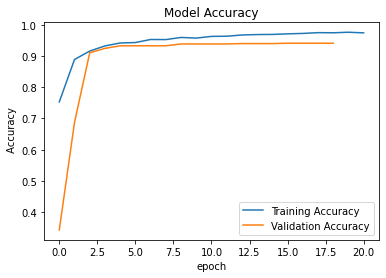}}
\caption{VGG-19 Training Accuracy and Validation Categorical Accuracy }
\label{fig}
\end{figure}

\subsubsection{DenseNet201}
Now, we start to work with another important convolutional neural network model
known as DenseNet 201. It is one of the latest neural network architectures available
that helped us to make our model even more reliable, consisting of 201 layers. We
kept the image size (224, 224, 3) during our model training procedure. Along with
this, to avoid over-training we have made the internal layer trainable to false and
added a dropout layer of value 0.5.
Next, we used a sequential model as our backbone architecture to pass the DenseNet
201 layers and make our custom model. We have added the Global Average Pooling
2D layer and a Dense layer with a size of 1024 to complete the process of designing
our custom DenseNet 201 model for improved performance, considering the fundamental DenseNet 201 model.
In addition, during training time we have calculated steps per epoch and the number
of epochs based on the available number of images and batch size count to keep our
model safe from overtraining. With all these careful steps, we have found 94.1
categorical validation accuracy during our training time.

\begin{figure}[htbp]
\centering{\includegraphics[scale=0.5]{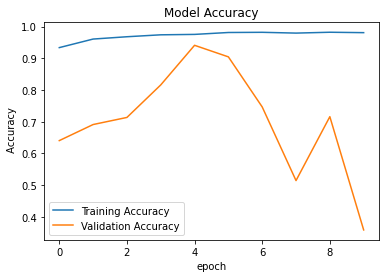}}
\caption{DenseNet201 Training Accuracy and Validation Categorical Accuracy}
\label{fig}
\end{figure}

\subsubsection{Inception V3}

Inception V3 is the third edition of Google’s Inception Convolutional Neural Network. We have used the latest pre-trained model for our disease detection system.
Inception V3 is a parallel processing architecture. The default input image size is
(299, 299, 3). However, we used (224, 224, 3) like our previously used model
VGG-19, DenseNet201.
We have used the Sequential Model as our backbone architecture during the implementation of the Inception V3 model. We have added the Global Average Pooling 2D
layer and a Dense layer size of 1024 during the development of our custom Inception
V3 model. We have added a dropout layer with a value of 0.5 to avoid over-fitting
problems.
Furthermore, we have calculated steps per epoch, which is 200 and the number of
epochs is 30 based on the image count of more than six thousand and batch size of
25. We have fine-tuned the training structure, keeping in mind our hardware limitations and avoiding over-training. With all these careful steps, we have found 94.5\%
of validation accuracy during our training time.

\begin{figure}[htbp]
\centering{\includegraphics[scale=0.5]{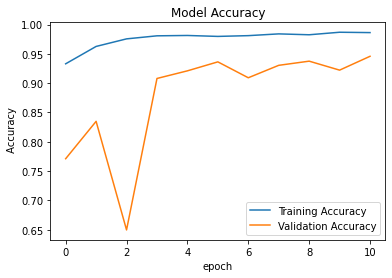}}
\caption{Inception V3 Training Accuracy and Validation Categorical Accuracy}
\label{fig}
\end{figure}

\subsubsection{SWIN Transformer}

SWIN Transformer is the CNN architecture with the original branch of the transformer-based learning approach. It is one of the most prominent architectures developed
by Microsoft. The full form of SWIN is Shifted Window. In this process, we can
reach the pixel-level image detail of an image.
This transformer learning technique divides the image into different patches before
sending it for training. Like the existing CNN model, the SWIN transformer is a
large encoder-decoder block that processes the input data.
SWIN Transformer is the general-purpose backbone of computer vision. The shifting
window technique brings astonishing efficiency by limiting self-attention computation to non-overlapping local windows while also allowing the cross-window connection. We have used a patch size of (2,2) and a number of attention heads of 8.
Moreover, we have used a window size of 7 with a shift size of 1. In our training
structure, we have maintained the image dimension is (224, 224, 3).
Furthermore, we have calculated steps per epoch that is 200 and the number of
epochs is 10 based on the image count of more than six thousand and batch size of 25. We have fine-tuned the training structure keeping in mind our hardware limitation and avoiding over-training. With all these careful steps, we have found 82.5\% of validation accuracy during our training time.

\subsubsection{Fusion Model}

We have individually trained and tested our discussed models. Now in order to
create the hybrid model. The hybrid model is the combination of transfer learning-based models (VGG-19, Inception V3, DenseNet 201) and the Transformer Model
(SWIN Transformer). We have developed this hybrid model.
We have used the ensemble technique to combine the entire model and build a single unique model that will work as the backbone of our disease detection system.

\begin{figure}[!ht]
\centering{\includegraphics[scale=0.5]{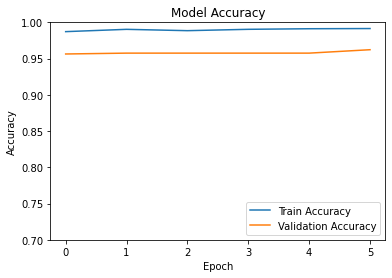}}
\caption{Training Accuracy and Validation Accuracy (Fusion Model)}
\label{fig}
\end{figure}
We are hopeful that this system that we have worked with COVID-19, Pneumonia,
and Normal X-ray image will also provide significant outcomes if this model is going
to use for any other detection system development.
Furthermore, we have used previous training configurations in order to maintain
the proper collaboration of the different models and ensure the best throughput
of the hybrid model. We have used (224, 224, 3) image size with a dropout layer
value of 0.5. Next, we have maintained a proper training structure consisting of the
number of epochs, steps per epoch, and batch size. With all these careful steps,
we have found 97.0\% of validation accuracy during our training time by combining
techniques that sum the weight of all four models. Next, we found 94.0\% of validation accuracy by combining techniques that average the weight of all four models.

In Fig 7, we have shown our trained dataset using Fusion Model. We have found training accuracy
99\% approximately and categorical validation accuracy of 96\% approximately. This indicates a slight over-fitting problem. However, we are working to improve this over-fitting problem in our upcoming research outcome.

\begin{figure}[!ht]
\centering{\includegraphics[scale=0.5]{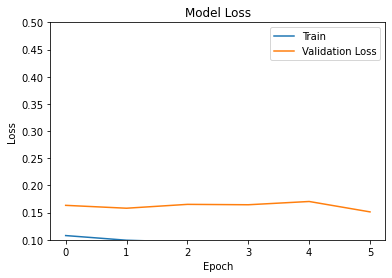}}
\caption{Training Loss and Validation Loss}
\label{fig}
\end{figure}

In Fig 8, We have found a training loss
of approximately less than 0.10\% and a categorical validation loss of less than 0.14\%
approximately using the Fusion Model. This low loss indicates that our model is overfitting despite using middle-layer trainable false and drop-out layer values of 0.5.

\subsection{Performance Analysis}
In Fig.9, we have shown our findings when we deployed our newly developed algorithm in a federated environment. We did not find an impressive outcome as we were facing hardware resource limitations. In a single run, the Federated environment consumed 35GB of RAM. 
\begin{figure}[htbp]
\subfigure{\includegraphics[width=0.5\textwidth]{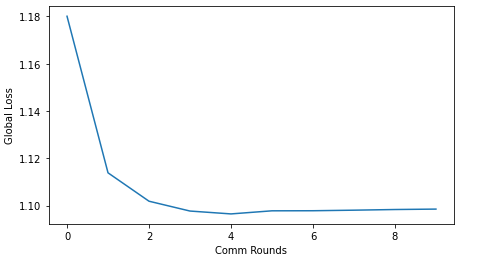}} 
\subfigure{\includegraphics[width=0.5\textwidth]{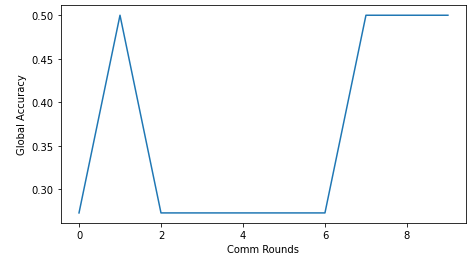}}
\caption{Federated Learning Based Outcome}
\end{figure}

\begin{figure}[htbp]

\subfigure{\includegraphics[width=0.48\textwidth]{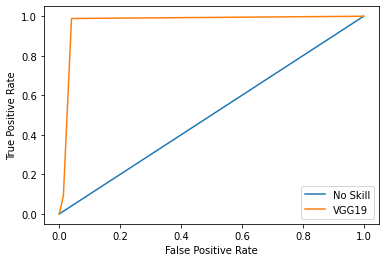}} 
\hspace{5mm}
\subfigure{\includegraphics[width=0.48\textwidth]{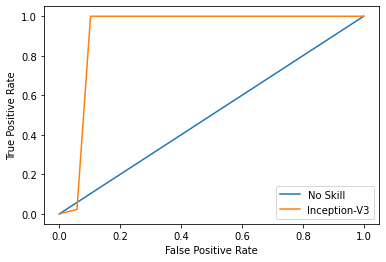}} \\
\subfigure{\includegraphics[width=0.48\textwidth]{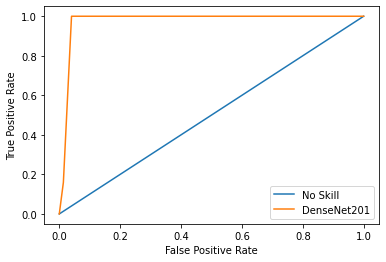}}
\hspace{5mm}
\subfigure{\includegraphics[width=0.48\textwidth]{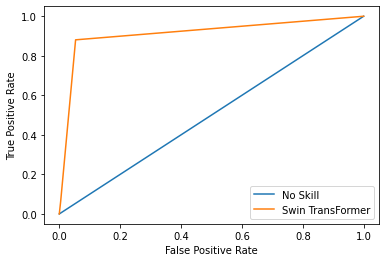}}
\caption{VGG19,Inception-V3 and DenseNet201 model's AUC-ROC outcomes}
\label{fig:foobar}
\end{figure}

In Fig. 10, the VGG19’s ROC-AUC curve shows a good Area Under the Curve than the
NoSkill straight line diagonal. Moreover, in Inception-V3 ROC-AUC curve shows less area under the curve than the
VGG-19 model. So the Inception-v3 is less accurate than the VGG-19 model. Furthermore, DenseNet 201 the value of AUC is also well, but initially, the area under the curve got shrinks but the ultimate area is good.

\begin{figure}[htbp]
\centering{\includegraphics[scale=0.5]{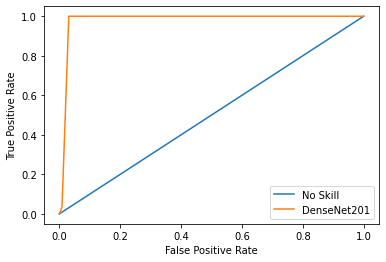}}
\caption{Fusion Model's AUC-ROC }
\label{fig}
\end{figure}
But in Fig. 11, This is the Fusion model’s AUC-ROC curve here after combining all the models we
get a good AUC-ROC curve with well AUC value than the other models individually.

\subsubsection{ Comparative Analysis}
In Fig. 12, Our comparative confusion matrix analysis has been shown between VGG19, Inception-v3, and DenseNet201. All individual models are predicting mostly True-Positive. But In Fig. 10 our Fusion model's True-Positive performance is better than each individual model.

\begin{figure}[htbp]
\centering
\subfigure{\includegraphics[width=0.45\textwidth]{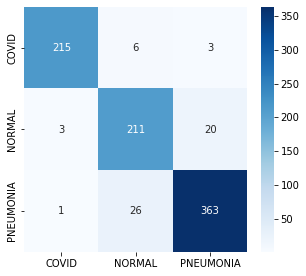}} 
\hspace{10mm}
\subfigure{\includegraphics[width=0.45\textwidth]{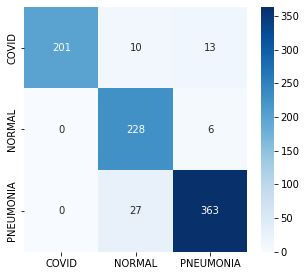}} \\
\vspace{10mm}
\subfigure{\includegraphics[width=0.45\textwidth]{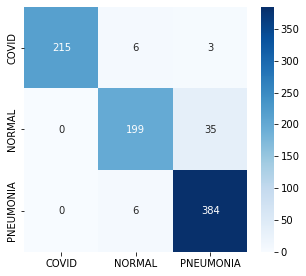}}
\hspace{10mm}
\subfigure{\includegraphics[width=0.45\textwidth]{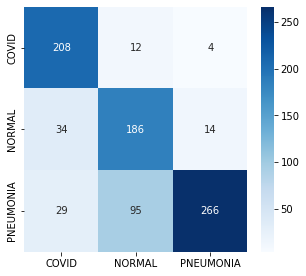}}\\
\caption{(a) VGG19 (top-left), (b) Inception-V3 (top-right), (c) DenseNet 201 (bottom-left) and (d) SWIN Transformer (bottom-right) model's Confusion Matrix  outcomes}
\label{fig:foobar}
\end{figure}

\begin{figure}[htbp]
\centering{\includegraphics[scale=0.5]{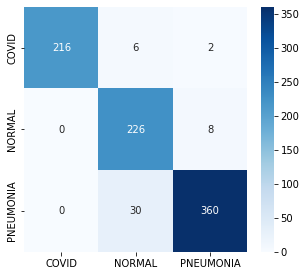}}
\caption{Fusion model's Confusion Matrix outcomes}
\label{fig}
\end{figure}

\begin{table}[!ht]
\caption{Model Comparison Table}
\scalebox{.85}{
\begin{tabular}{|cccc|}
\hline
\multicolumn{4}{|c|}{\textbf{Model Comparison}}                                                                                                                                                                                                          \\ \hline
\multicolumn{1}{|c|}{\textbf{Classifier}}             & \multicolumn{1}{c|}{\textbf{\begin{tabular}[c]{@{}c@{}}Training Time (s)\\ (approx.)\end{tabular}}} & \multicolumn{1}{c|}{\textbf{\begin{tabular}[c]{@{}c@{}}Testing Time (s) \\ (approx.)\end{tabular}}} & \textbf{Accuracy (\%)} \\ \hline
\multicolumn{1}{|c|}{\textbf{VGG-19}}                 & \multicolumn{1}{c|}{\textbf{14440}}                                                                 & \multicolumn{1}{c|}{\textbf{4}}                                                                     & \textbf{94.4}          \\ \hline
\multicolumn{1}{|c|}{\textbf{Inception V3}}           & \multicolumn{1}{c|}{\textbf{15200}}                                                                 & \multicolumn{1}{c|}{\textbf{2}}                                                                     & \textbf{94.5}          \\ \hline
\multicolumn{1}{|c|}{\textbf{DenseNet 201}}           & \multicolumn{1}{c|}{\textbf{18120}}                                                                 & \multicolumn{1}{c|}{\textbf{2}}                                                                     & \textbf{94.1}          \\ \hline
\multicolumn{1}{|c|}{\textbf{SWIN Transformer}}       & \multicolumn{1}{c|}{\textbf{25650}}                                                                 & \multicolumn{1}{c|}{\textbf{4}}                                                                     & \textbf{82.5}          \\ \hline
\multicolumn{1}{|c|}{\textbf{Fusion Model (Sum)}}     & \multicolumn{1}{c|}{\textbf{24122}}                                                                 & \multicolumn{1}{c|}{\textbf{3}}                                                                     & \textbf{96.24}         \\ \hline
\multicolumn{1}{|c|}{\textbf{Fusion Model (Average)}} & \multicolumn{1}{c|}{\textbf{21600}}                                                                 & \multicolumn{1}{c|}{\textbf{2}}                                                                     & \textbf{94}            \\ \hline

\end{tabular}
}
\label{tab: Model Comparison Table}

\end{table}
We have observed different performance criteria. Among them, we have found the Confusion Matrix-based test outcome that gives us an idea about the performance comparison.
\section{Conclusion}

In this research, we have provided brief explanations on \textbf{A Hybrid FL-Enabled Ensemble Approach For Lung Disease Diagnosis Leveraging a Combination of SWIN Transformer and CNN.} We have used a combined model of transfer learning and transformer learning known as shifted window transformer to make our model reliable. In the future, we want to work on the concept drift of federated learning to address the limitation of federated learning. In addition, we want to improve our Algorithm to make it more efficient and reliable during the analysis process of patient data. We want to improve our individual algorithm's efficiency and use better hardware to train our models that will be efficient and reliable. We also want to analyze our program's time and space complexity. In addition, we want to analyze our dataset from a statistical aspect as well. Lastly, we want to use our developed system in different disease detection processes to contribute to the medical sector. However, technology will always work as assistance to medical treatment but will be limited because of the variance of diseases and treatment processes.

%
%

\begin{thebibliography}{8}

\bibitem{ref_article1}
Kassania, S. H., Kassanib, P. H., Wesolowskic, M. J., Schneidera, K. A., \& Detersa, R. (2021).
Automatic detection of coronavirus disease (COVID-19) in X-ray and CT images: A machine learning based approach.
Biocybernetics and Biomedical Engineering, 41(3), 867--879.

\bibitem{ref_lncs1}
Hemdan, E. E. D., Shouman, M. A., \& Karar, M. E. (2020).
Covidx-net: A framework of deep learning classifiers to diagnose COVID-19 in X-ray images.
arXiv preprint arXiv:2003.11055.

\bibitem{ref_book1}
Minaee, S., Kafieh, R., Sonka, M., Yazdani, S., \& Soufi, G. J. (2020).
Deep-COVID: Predicting COVID-19 from chest X-ray images using deep transfer learning.
Medical Image Analysis, 65, 101794.

\bibitem{ref_proc1}
Jiang, J., \& Lin, S. (2021).
Covid-19 detection in chest X-ray images using SWIN-transformer and transformer in transformer.
arXiv preprint arXiv:2110.08427.

\bibitem{ref_sthardnet}
Gu, Y., Piao, Z., \& Yoo, S. J. (2022).
STHarDNet: SWIN Transformer with HarDNet for MRI segmentation.
Applied Sciences, 12(1), 468.

\bibitem{ref_fl_challenges}
Li, T., Sahu, A. K., Talwalkar, A., \& Smith, V. (2020).
Federated learning: Challenges, methods, and future directions.
IEEE Signal Processing Magazine, 37(3), 50--60.

\bibitem{ref_variational_fl}
Corinzia, L., Beuret, A., \& Buhmann, J. M. (2019).
Variational federated multi-task learning.
arXiv preprint arXiv:1906.06268.

\bibitem{ref_multi_task}
Smith, V., Chiang, C. K., Sanjabi, M., \& Talwalkar, A. S. (2017).
Federated multi-task learning.
Advances in Neural Information Processing Systems, 30.


\end{thebibliography}
%

\end{document}